\documentclass[runningheads]{llncs}
\usepackage[T1]{fontenc}
\usepackage{graphicx,verbatim}
\usepackage{booktabs}
\usepackage{multirow}
\usepackage{xcolor}
\usepackage{makecell}
\usepackage{pifont}
\usepackage{array}
\usepackage{amsmath,amssymb}
\usepackage{enumitem}
\usepackage{tabularx}
\usepackage{capt-of}
\newcommand{\ours}{HERO}

\begin{document}

\title{HERO: Hypothesis-Driven Evidence Retrieval from Omics for Multi-Task Breast Cancer Analysis}
\titlerunning{HERO: Hypothesis-Driven Evidence Retrieval from Omics}

\author{Xiangyu Li \and Ran Su\thanks{Corresponding author}}
\authorrunning{X. Li and R. Su}
\institute{College of Intelligence and Computing, Tianjin University, Tianjin 300072, China\\
    \email{xiangyuli@tju.edu.cn, ran.su@tju.edu.cn}}
\maketitle

\begin{abstract}
Matched multi-omics can improve WSI-based biomarker and prognosis prediction, but most existing pipelines use omics as a parallel feature stream or textual context rather than as an explicit retrieval constraint.
\ours{} asks whether observed omics can be a testable morphology hypothesis: a sparse pathway-to-morphology prior maps DNA methylation and miRNA into a $K$-dimensional intent vector $\mathbf{m}$ ($K{=}16$), TF-IDF retrieval over structured 10$\times$ captions selects endpoint-relevant regions, and a cosine gate $c{=}\cos(\mathbf{m},\mathbf{v})$ triggers deterministic deficit-driven repair when $c{<}\tau_c$.
This closed-loop design bounds VLM calls, reduces reliance on embedding-based semantic matching, and makes every retrieval and verification step lexically auditable.
On TCGA-BRCA (930 WSIs, patient-level 5-fold CV), \ours{} sets new state-of-the-art across ER, PR, HER2, subtype, and risk prediction, outperforming both multimodal fusion and VLM-based baselines.
\end{abstract}

\section{Introduction}\label{sec:intro}

Breast cancer diagnosis requires joint evidence from H\&E whole-slide images (WSIs) and molecular biomarkers (ER/PR/HER2, molecular subtype, prognostic signatures)~\cite{xu2024whole,ding2025titan}.
A clinically useful computational system should not only predict endpoints but also produce an auditable evidence trail: which tissue regions were examined, what morphology was observed, and whether the observations support the molecular conclusion.

\begin{figure}[t]
    \centering
    \includegraphics[width=0.7\textwidth]{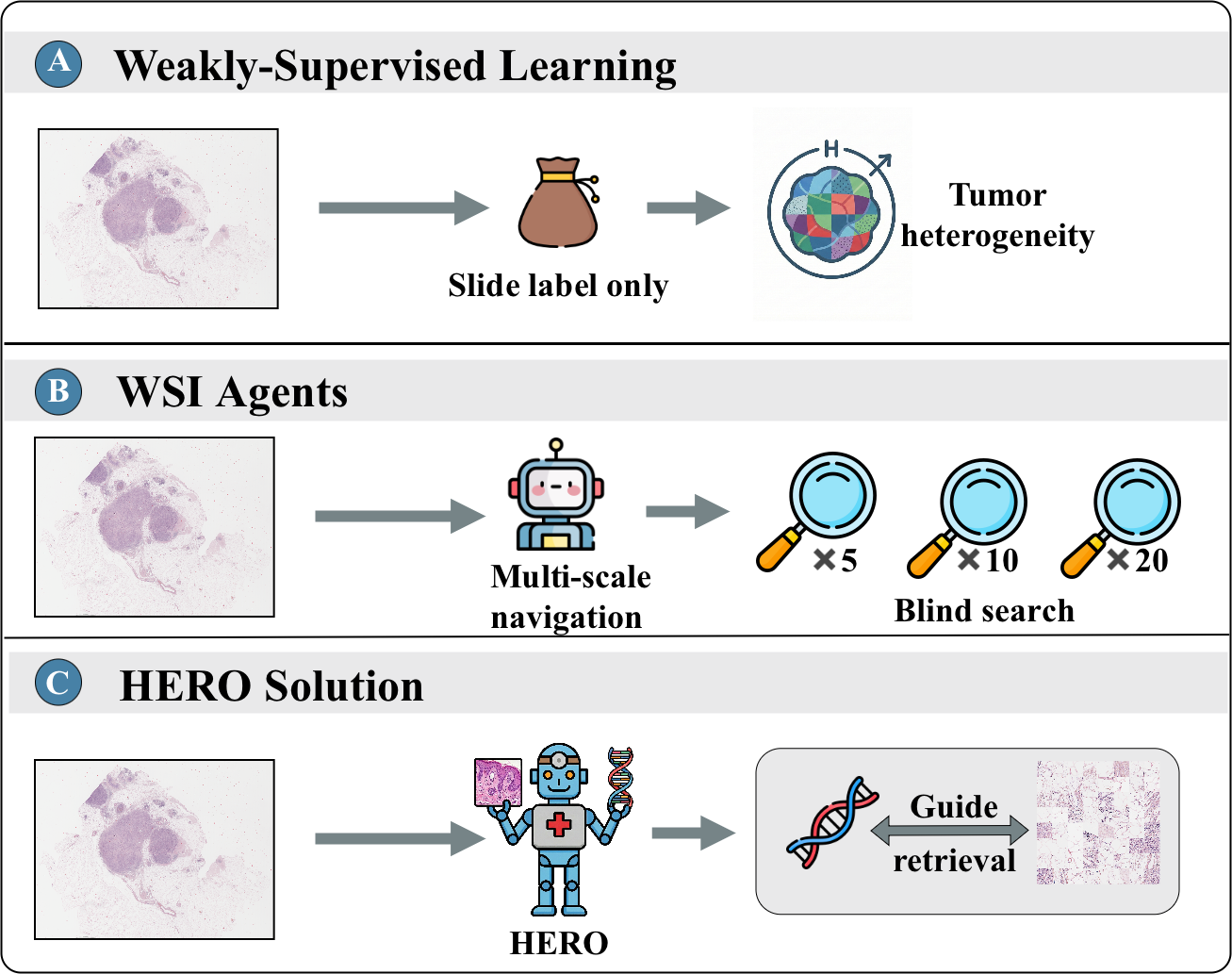}
    \caption{(A)~MIL relies on slide-level labels; attention may highlight non-diagnostic regions under intratumoral heterogeneity.
    (B)~VLM-based WSI readers can suffer retrieval bias toward visually salient regions.
    (C)~\ours{} uses omics-derived intent to control retrieval and a consistency gate to verify molecular--visual alignment.}
    \label{fig:motivation}
\end{figure}

The central technical challenge is \emph{endpoint-aware evidence retrieval} on gigapixel WSIs under intratumoral heterogeneity.
Three families of methods address this problem, each with a specific limitation.
(i)~Weakly supervised MIL~\cite{ilse2018attention_icml,shao2021transmil_neurips} aggregates patches under slide-level labels; attention is post-hoc and does not verify evidence sufficiency.
(ii)~VLM-based WSI readers and multi-agent systems~\cite{Liang_2025_ICCV,Chen_2025_CVPR,kim2024mdagents,tang-etal-2024-medagents} make reading explicit, but often rely on prompt-level guidance rather than a fixed, molecularly conditioned retrieval prior, which can cause \emph{retrieval bias} toward visually salient but endpoint-irrelevant regions.
(iii)~Multimodal WSI--omics fusion methods~\cite{chen2021mcat,jaume2024survpath} and omics-as-context reasoning systems~\cite{fallahpour2025medrax,li-etal-2024-mmedagent} treat omics as a parallel feature stream or appended text; omics does not control \emph{which} regions are retrieved, nor is it used to \emph{verify} retrieval completeness.

A complementary literature predicts molecular signals directly from H\&E morphology~\cite{arslan2024pancancer,mondol2023hist2rna,ekholm2024proliferation,liu2024wsi2rppa,wang2024screenthemall,wu2025virtualihc}, showing that morphology encodes molecular information.
\ours{} operates in the reverse direction: it uses \emph{observed} omics to control retrieval and verify evidence completeness, which is the step most existing pipelines leave implicit.
We also note interpretable fusion with spatial transcriptomics~\cite{liu2026pathospatial}.

We introduce \ours{}, a four-stage pipeline (Fig.~\ref{fig:motivation}) with three contributions:
\begin{enumerate}[nosep,leftmargin=*]
    \item Omics-guided retrieval: a sparse pathway-to-morphology prior converts DNA methylation and miRNA into a $K$-dimensional intent vector $\mathbf{m}$ and keyword set $\mathcal{Q}$, enabling endpoint-aware patch selection via TF-IDF over structured captions.
    \item Consistency gate with deficit-driven repair: a cosine gate $c{=}\cos(\mathbf{m},\mathbf{v})$ detects molecular--visual mismatch; when $c{<}\tau_c$, deterministic re-ranking retrieves patches that cover under-supported morphology axes, bounding compute to one repair round.
    \item Robustness analysis: we evaluate sensitivity to $K$, $\tau_c$, and perturbations of the prior $\mathbf{W}$, providing evidence that the mapping encodes informative biological structure rather than arbitrary assignments.
\end{enumerate}

\section{Method}\label{sec:method}

\noindent\textbf{Problem formulation.}
Given a WSI $I$ with matched DNA methylation $\mathbf{x}_{\mathrm{dna}}$ and miRNA $\mathbf{x}_{\mathrm{mirna}}$, we predict ER/PR/HER2, subtype, and risk while producing an auditable evidence chain.
A shared $K$-dimensional evidence space ($K{=}16$) contains a molecular intent vector $\mathbf{m}\in\mathbb{R}^K$, a visual evidence vector $\mathbf{v}\in\mathbb{R}^K$, and a consistency score $c{=}\cos(\mathbf{m},\mathbf{v})$.
We use TF-IDF and word-boundary checklist matching throughout: pathology descriptors that are semantically similar can be diagnostically distinct (e.g., ``high-grade nuclear atypia'' vs.\ ``reactive atypia''), and embedding-based retrieval can conflate such terms.
Structured prompts elicit standardized terminology for reliable lexical scoring.

\subsection{Stage 1: Omics to intent and keywords}\label{sec:s1}

Given $\mathbf{x}_{\mathrm{dna}}\in\mathbb{R}^{F_d}$ and $\mathbf{x}_{\mathrm{mirna}}\in\mathbb{R}^{F_m}$, this stage produces an intent vector $\mathbf{m}\in\mathbb{R}^K$, a keyword set $\mathcal{Q}$, and a molecular narrative.

\paragraph{Pathway scoring.}
We z-normalize each omics vector and compute pathway scores via pre-computed binary masks $\mathbf{M}_{\mathrm{dna}}\in\mathbb{R}^{P\times F_d}$, $\mathbf{M}_{\mathrm{mirna}}\in\mathbb{R}^{P\times F_m}$ derived from MSigDB Hallmark gene sets \cite{liberzon2015hallmark} ($P{=}50$) and validated miRNA-target interactions (e.g., miRTarBase) \cite{chou2018mirtarbase}. For pathway $p$:
\begin{equation}
\begin{aligned}
s^{\mathrm{dna}}_p
&= -\frac{\sum_{j=1}^{F_d}M^{\mathrm{dna}}_{p,j}z(x^{\mathrm{dna}}_j)}
{\sum_{j=1}^{F_d}M^{\mathrm{dna}}_{p,j}+\epsilon},\\
s^{\mathrm{mirna}}_p
&= -\frac{\sum_{j=1}^{F_m}M^{\mathrm{mirna}}_{p,j}z(x^{\mathrm{mirna}}_j)}
{\sum_{j=1}^{F_m}M^{\mathrm{mirna}}_{p,j}+\epsilon},\qquad
s_p=\frac{s^{\mathrm{dna}}_p+s^{\mathrm{mirna}}_p}{2}.
\end{aligned}
\end{equation}
The denominator is a pathway-wise average over assigned features, not a global matrix norm; the negation maps hypermethylation and miRNA-mediated repression to pathway down-regulation scores.
This linear scoring provides interpretability, resists overfitting on a 930-WSI cohort, and is regularized by the sparse prior $\mathbf{W}$ below.
\begin{figure}[t]
    \centering
    \includegraphics[width=\textwidth]{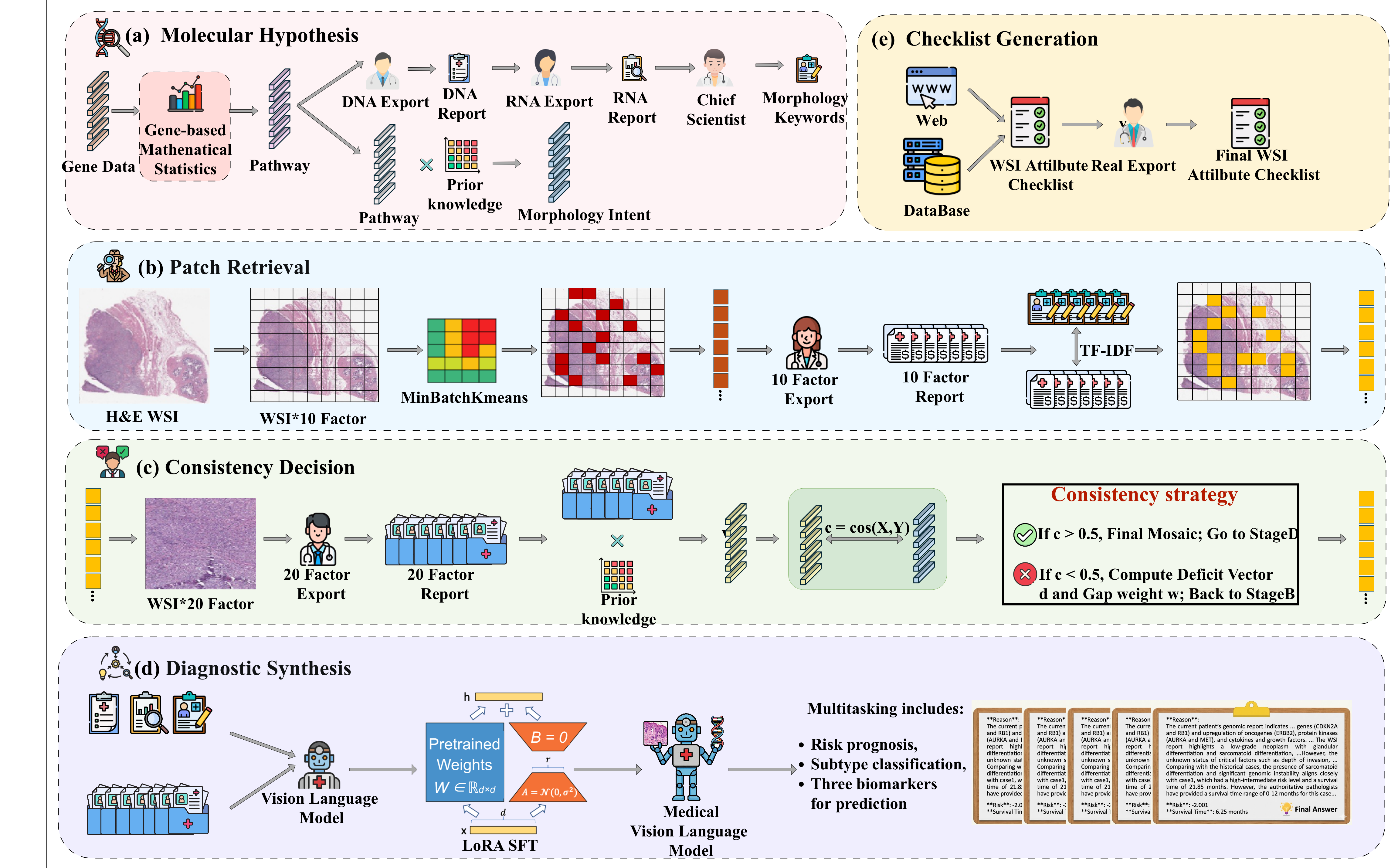}
    \caption{Overview of \ours{}.
    (a)~Stage~1: omics$\rightarrow$intent via pathway scoring and committee;
    (b)~Stage~2: 10$\times$ representative mining and TF-IDF retrieval;
    (c)~Stage~3: 20$\times$ consistency gate and deficit-driven repair;
    (d)~Stage~4: LoRA-tuned VLM diagnosis;
    (e)~morphology axis checklist shared across stages.}
    \label{fig:framework}
\end{figure}
\paragraph{Morphology axes, checklist, and prior $\mathbf{W}$.}
We define $K{=}16$ axes: mitotic activity, nuclear pleomorphism, tubule formation, necrosis, lymphocytic infiltrate, stromal desmoplasia, solid growth, cribriform pattern, mucin production, vascular invasion, nuclear grade, apoptotic bodies, inflammatory response, cell density, architectural disorder, and calcification.
Each axis $k$ has a checklist $\mathcal{C}_k$ of 10--30 descriptors compiled from breast pathology terminology; synonyms are merged and ambiguous terms removed. The checklist is fixed before cross-validation and used unchanged in all folds.
Matching uses case-insensitive word boundaries.
Instead of unconstrained black-box mapping, we construct a signed, weighted sparse prior $\mathbf{W}\in\mathbb{R}^{K\times P}$ guided by WHO terminology and biological associations (E2F/MYC $\rightarrow$ proliferation; EMT/TGF-$\beta$ $\rightarrow$ stroma; hypoxia $\rightarrow$ necrosis); rows are $\ell_1$-normalized and fixed before cross-validation.
This fixed prior does not by itself rule out shortcut learning. Instead, it makes the omics-to-morphology mapping explicit, fold-invariant, and auditable; $\mathbf{W}$ is never optimized with endpoint labels.
$\mathbf{W}$ serves as an auditable prior and explicit regularizer: $\mathbf{W}$-shuffle degrades HER2 by 0.033 and Risk by 0.032, while noise/drop cause $\le$0.009 degradation (Table~\ref{tab:robustness}).

\paragraph{Intent vector.}
\begin{equation}
    \mathbf{m} = \ell_2\text{-normalize}\!\left(\mathrm{ReLU}(\mathbf{W}\mathbf{s})\right)\in\mathbb{R}^{K}.
\end{equation}

\paragraph{Committee module.}
A role-based chain of three LLM calls returns a structured record with up/down-regulated pathways, morphology axes, a molecular narrative, and a 15--40 term keyword set $\mathcal{Q}$ after synonym merging and removal of ambiguous descriptors. The keyword set is used only for Stage~2 lexical retrieval.

\subsection{Stage 2: 10$\times$ retrieval (300$\rightarrow$Top-64)}\label{sec:s2}

\paragraph{Representative mining ($N_r{=}300$).}
Tissue patches (256$\times$256\,px at 20$\times$) are encoded with $\ell_2$-normalized UNI~\cite{chen2024uni} embeddings (1024-dim).
MiniBatchKMeans ($k{=}N_r{=}300$, batch size 1024) clusters the embeddings; the patch closest to each centroid is selected as a representative, ensuring coverage of the slide's global morphological diversity.

\paragraph{10$\times$ captioning and TF-IDF ranking.}
Each representative $p_i$ is revisited at 10$\times$ and captioned by Qwen2.5-VL-7B~\cite{qwen2vl} with a structured prompt covering tumor architecture, stromal composition, inflammatory infiltrate, and necrosis, producing $\mathrm{cap}_i^{10}$.
A TF-IDF vectorizer $\phi(\cdot)$ is fit on the 300-caption corpus; each representative is scored against the keyword set $\mathcal{Q}$:
\begin{equation}
\mathrm{score}_{10}(p_i)=\cos\!\big(\phi(\mathcal{Q}),\;\phi(\mathrm{cap}_i^{10})\big).
\end{equation}
The vectorizer is fit separately per slide, so IDF is a within-slide rarity weight rather than a cohort-level learned parameter; no endpoint label is used.
The Top-64 by $\mathrm{score}_{10}$ form the evidence set for Stage~3, corresponding to an $8{\times}8$ mosaic (single VLM forward pass).

\subsection{Stage 3: 20$\times$ consistency gate and repair}\label{sec:s3}

\paragraph{Mosaic construction and captioning.}
The Top-64 regions are cropped at 20$\times$ (256$\times$256\,px), arranged into a coordinate-labeled $8{\times}8$ mosaic, and captioned with the same structured prompt, now requesting cytologic features such as nuclear grade, mitoses, chromatin pattern, and cell-level morphology.

\paragraph{Visual evidence vector $\mathbf{v}$.}
For caption $i$ and axis $k$, $u_{ik}=1$ if any checklist term in $\mathcal{C}_k$ appears in $\mathrm{cap}_i^{20}$ under word-boundary matching. We aggregate $v_k=\sum_{i=1}^{64}u_{ik}$ and $\ell_2$-normalize $\mathbf{v}$; if all $v_k{=}0$, $\mathbf{v}{=}\mathbf{0}$.

\paragraph{Consistency gate and repair.}
The initial Top-64 mosaic is captioned once at 20$\times$. We compute:
\begin{align}
c &= \cos(\mathbf{m},\mathbf{v})
=\frac{\mathbf{m}^{\top}\mathbf{v}}{\|\mathbf{m}\|_2\|\mathbf{v}\|_2}, \\
\mathbf{w}' &= \mathrm{softmax}\!\left(\mathrm{ReLU}(\mathbf{m}-\mathbf{v})/\tau\right),\quad \tau{=}0.3,\\
\mathrm{score}_{\mathrm{repair}}(p_i)
&= \mathrm{score}_{10}(p_i)+{\mathbf{w}'}^{\top}\mathbf{a}_i .
\end{align}
We set $c{=}0$ when $\mathbf{v}{=}\mathbf{0}$ and accept the mosaic if $c\ge\tau_c$ ($\tau_c{=}0.5$).
If $c<\tau_c$, the deficit $\mathbf{m}-\mathbf{v}$ identifies under-supported axes. Each unselected representative receives an axis-hit vector $\mathbf{a}_i\in\{0,1\}^K$ from its 10$\times$ caption; the top-32 repair candidates are merged with the original 64, the final Top-64 is rebuilt from this 96-patch pool, and one additional 20$\times$ captioning pass is performed. No further repair round is performed.

\subsection{Stage 4: Diagnosis module}\label{sec:s4}

The final 20$\times$ mosaic, molecular narrative, and consistency summary (including $c$ and top deficit axes when repair was triggered) are fed into Qwen2.5-VL-7B~\cite{qwen2vl} with LoRA~\cite{hu2022lora_iclr} (rank 64, $\alpha{=}128$, \texttt{q/k/v/o\_proj}; backbone frozen).
The molecular narrative is textual context, not an additional learned omics feature; the consistency summary reports $c$ and whether visual evidence already supports the molecular hypothesis or required repair.
The model outputs JSON predictions for ER, PR, HER2, subtype, and risk.
Training uses teacher forcing with cross-entropy loss on the structured JSON tokens; at inference, greedy decoding is used.
Only Stage~4 is tuned because Stages~1--3 are intended to remain fixed retrieval and verification modules; tuning the retrieval prior or repair rule with endpoint labels would make it harder to isolate evidence selection from final diagnosis.
All adaptivity is deterministic (gate + re-ranking); the VLM budget per slide is 300 (10$\times$) + 64 (20$\times$) + at most 32 repair (20$\times$, triggered on 37\% of slides).
$\{\mathcal{C}_k\}$ and $\mathbf{W}$ are fixed across folds.

\section{Experiments}\label{sec:exp}
\subsection{Setup}\label{sec:setup}
We evaluate on TCGA-BRCA (930 WSIs, 878 patients) with matched DNA methylation (14{,}115 CpGs) and miRNA-seq (315 miRNAs) for ER/PR/HER2, subtype (5-class), and survival risk, using patient-level 5-fold stratified cross-validation (seed 42).
Labels for ER, PR, HER2, clinical subtype, and risk are from TCGA clinical annotations; missing or equivocal receptor status is excluded per endpoint. These endpoint labels are not produced by \ours{} from its input methylation and miRNA vectors.
Only the Stage~4 model is LoRA-tuned; UNI features (1024-dim, 20$\times$) are pre-extracted.
Baselines: omics-only (SNN~\cite{klambauer2017snn_neurips}, SNNTrans~\cite{zhou2024ccl_tmi}), WSI-only MIL (ABMIL~\cite{ilse2018attention_icml}, TransMIL~\cite{shao2021transmil_neurips}), multimodal fusion (MCAT~\cite{chen2021mcat}, MOTCat~\cite{xu2023motcat}, SurvPath~\cite{jaume2024survpath}), VLM/agent (GPT-4o, MDAgent~\cite{kim2024mdagents}).
For fair comparison, all multimodal baselines were implemented to consume the exact same concatenated DNA methylation and miRNA feature vectors as our method.
This unified protocol controls input modality across methods, but may differ from the originally optimized configuration of some baselines; results are interpreted under a common methylation+miRNA setting rather than as a claim that each baseline is globally optimized.
Inference: 57\,s/WSI on 2$\times$A100, 22\,GB peak; the gate triggers on 37\% of test slides.

\subsection{Comparative results}\label{sec:compare}
Table~\ref{tab:main} reports results across all endpoints.
\ours{} obtains the highest scores on every metric.
The largest improvements over the best fusion baseline (SurvPath) are on HER2 (+0.050 AUC) and Risk (+0.030 C-index), the two endpoints where diagnostically relevant morphology is sparse and heterogeneous, making retrieval quality a bottleneck.

To isolate the contribution of omics-guided retrieval from VLM capacity, we evaluate two controlled baselines that use the \emph{same} LoRA-tuned Qwen2.5-VL-7B diagnosis head as \ours{} but replace the retrieval mechanism:
VLM + TransMIL Top-64 (attention-selected patches) and VLM + Random Top-64 (uniformly sampled patches).
\ours{} improves over VLM + TransMIL Top-64 by +0.090 on HER2 and +0.070 on Risk, indicating that gains originate from molecular-guided retrieval and verification rather than model scale.
VLM + Random Top-64 is much lower (HER2 0.625, Risk 0.565), confirming that a 7B VLM without informed retrieval is insufficient.

\begin{table}[t]
\centering
\caption{Results on TCGA-BRCA (patient-level 5-fold CV, mean$\pm$std).
AUC for ER/PR/HER2, macro AUC for Subtype, C-index for Risk.
``*'': our reimplementation.
VLM baselines use the same LoRA-tuned Qwen2.5-VL-7B as \ours{} with Random or TransMIL-selected Top-64 evidence.}
\label{tab:main}
\scriptsize
\setlength{\tabcolsep}{1.8pt}
\newcolumntype{M}{>{\raggedright\arraybackslash}p{2.8cm}}
\begin{tabular}{Mccccc}
\toprule
Method & ER (AUC) & PR (AUC) & HER2 (AUC) & Subtype (AUC) & Risk (C-idx) \\
\midrule
SNN*~\cite{klambauer2017snn_neurips} & 0.952{\scriptsize$\pm$0.021} & 0.885{\scriptsize$\pm$0.035} & 0.725{\scriptsize$\pm$0.045} & 0.958{\scriptsize$\pm$0.015} & 0.620{\scriptsize$\pm$0.045} \\
SNNTrans*~\cite{zhou2024ccl_tmi} & 0.968{\scriptsize$\pm$0.018} & 0.912{\scriptsize$\pm$0.028} & 0.765{\scriptsize$\pm$0.038} & 0.970{\scriptsize$\pm$0.012} & 0.645{\scriptsize$\pm$0.038} \\
ABMIL~\cite{ilse2018attention_icml} & 0.885{\scriptsize$\pm$0.032} & 0.755{\scriptsize$\pm$0.045} & 0.655{\scriptsize$\pm$0.052} & 0.955{\scriptsize$\pm$0.022} & 0.590{\scriptsize$\pm$0.052} \\
TransMIL~\cite{shao2021transmil_neurips} & 0.901{\scriptsize$\pm$0.028} & 0.763{\scriptsize$\pm$0.041} & 0.687{\scriptsize$\pm$0.048} & 0.972{\scriptsize$\pm$0.015} & 0.615{\scriptsize$\pm$0.048} \\
MCAT*~\cite{chen2021mcat} & 0.975{\scriptsize$\pm$0.022} & 0.935{\scriptsize$\pm$0.031} & 0.768{\scriptsize$\pm$0.042} & 0.975{\scriptsize$\pm$0.018} & 0.665{\scriptsize$\pm$0.042} \\
MOTCat*~\cite{xu2023motcat} & 0.982{\scriptsize$\pm$0.019} & 0.948{\scriptsize$\pm$0.025} & 0.795{\scriptsize$\pm$0.036} & 0.980{\scriptsize$\pm$0.014} & 0.685{\scriptsize$\pm$0.035} \\
SurvPath*~\cite{jaume2024survpath} & 0.991{\scriptsize$\pm$0.015} & 0.957{\scriptsize$\pm$0.022} & 0.815{\scriptsize$\pm$0.032} & 0.985{\scriptsize$\pm$0.012} & 0.705{\scriptsize$\pm$0.030} \\
GPT-4o* & 0.945{\scriptsize$\pm$0.035} & 0.880{\scriptsize$\pm$0.045} & 0.710{\scriptsize$\pm$0.055} & 0.950{\scriptsize$\pm$0.025} & 0.610{\scriptsize$\pm$0.065} \\
MDAgent*~\cite{kim2024mdagents} & 0.965{\scriptsize$\pm$0.025} & 0.915{\scriptsize$\pm$0.038} & 0.755{\scriptsize$\pm$0.048} & 0.972{\scriptsize$\pm$0.018} & 0.640{\scriptsize$\pm$0.050} \\
\midrule
VLM + Rand.\ Top-64 & 0.845{\scriptsize$\pm$0.045} & 0.720{\scriptsize$\pm$0.055} & 0.625{\scriptsize$\pm$0.060} & 0.865{\scriptsize$\pm$0.035} & 0.565{\scriptsize$\pm$0.050} \\
VLM + TransMIL Top-64 & 0.962{\scriptsize$\pm$0.020} & 0.905{\scriptsize$\pm$0.030} & 0.775{\scriptsize$\pm$0.040} & 0.978{\scriptsize$\pm$0.015} & 0.665{\scriptsize$\pm$0.035} \\
\ours{} (Ours) & \textbf{0.994{\scriptsize$\pm$0.010}} & \textbf{0.978{\scriptsize$\pm$0.015}} & \textbf{0.865{\scriptsize$\pm$0.022}} & \textbf{0.995{\scriptsize$\pm$0.008}} & \textbf{0.735{\scriptsize$\pm$0.025}} \\
\bottomrule
\end{tabular}
\end{table}

\begin{table}[t]
\centering
\begin{minipage}[t]{0.48\textwidth}
\centering
\captionof{table}{Ablation on HER2 and Risk (mean$\pm$std).
Gate-only computes $c$ without repair; Random repair adds 32 random patches.}
\label{tab:ablation_hr}
\scriptsize
\setlength{\tabcolsep}{2pt}
\resizebox{\linewidth}{!}{%
\begin{tabular}{lcc}
\toprule
Setting & HER2 (AUC) & Risk (C-idx) \\
\midrule
Visual-only Top-64 & 0.758$\pm$0.048 & 0.645$\pm$0.050 \\
Text-only retrieval & 0.835$\pm$0.030 & 0.690$\pm$0.035 \\
\midrule
w/o consistency gate & 0.822$\pm$0.036 & 0.695$\pm$0.038 \\
Gate-only & 0.828$\pm$0.032 & 0.702$\pm$0.032 \\
Random repair & 0.830$\pm$0.040 & 0.700$\pm$0.045 \\
\midrule
\ours{} (full) & \textbf{0.865$\pm$0.022} & \textbf{0.735$\pm$0.025} \\
\bottomrule
\end{tabular}}
\end{minipage}\hfill
\begin{minipage}[t]{0.48\textwidth}
\centering
\captionof{table}{Sensitivity to $K$, $\tau_c$, and $\mathbf{W}$ perturbations (mean$\pm$std; HER2 AUC / Risk C-index).}
\label{tab:robustness}
\scriptsize
\setlength{\tabcolsep}{2pt}
\resizebox{\linewidth}{!}{%
\begin{tabular}{lcc}
\toprule
Setting & HER2 (AUC) & Risk (C-idx) \\
\midrule
$K{=}8$ & 0.852$\pm$0.026 & 0.724$\pm$0.028 \\
$K{=}16$ & 0.865$\pm$0.022 & 0.735$\pm$0.025 \\
$K{=}32$ & \textbf{0.867$\pm$0.023} & \textbf{0.736$\pm$0.026} \\
\midrule
$\tau_c{=}0.3$ & 0.858$\pm$0.024 & 0.728$\pm$0.027 \\
$\tau_c{=}0.5$ & \textbf{0.865$\pm$0.022} & \textbf{0.735$\pm$0.025} \\
$\tau_c{=}0.7$ & 0.861$\pm$0.025 & 0.731$\pm$0.026 \\
\midrule
$\mathbf{W}$ (ours) & \textbf{0.865$\pm$0.022} & \textbf{0.735$\pm$0.025} \\
$\mathbf{W}$-noise & 0.860$\pm$0.025 & 0.732$\pm$0.028 \\
$\mathbf{W}$-drop & 0.856$\pm$0.027 & 0.729$\pm$0.029 \\
$\mathbf{W}$-shuffle & 0.832$\pm$0.035 & 0.703$\pm$0.038 \\
\bottomrule
\end{tabular}}
\end{minipage}
\end{table}

\begin{figure}[t]
    \centering
    \includegraphics[width=\textwidth]{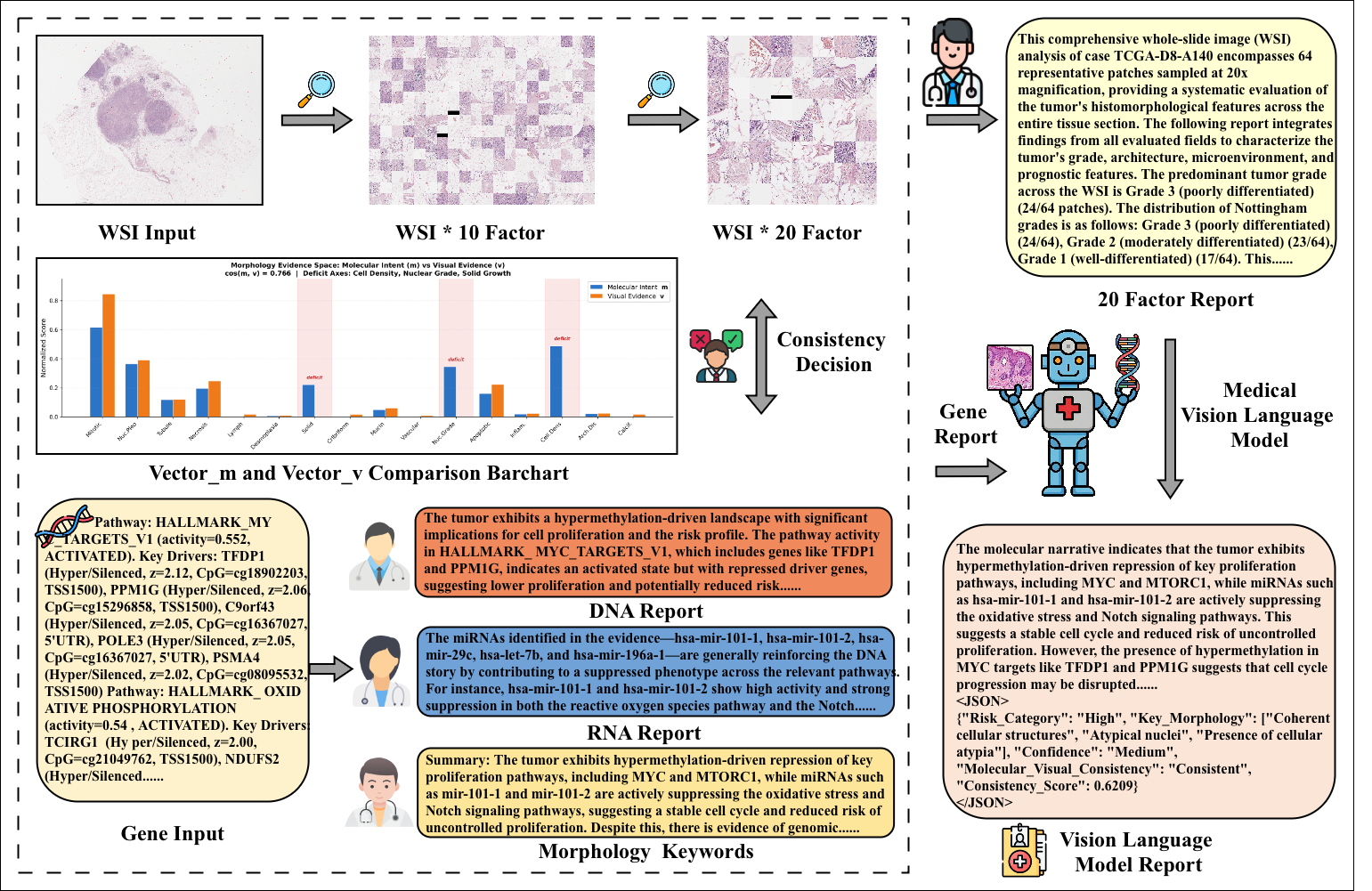}
    \caption{Case-level evidence chain.
    Omics$\rightarrow$intent $\mathbf{m}$ guides initial retrieval; captions yield $\mathbf{v}$ and $c{=}\cos(\mathbf{m},\mathbf{v})$.
    When $c{<}\tau_c$, repair candidates rebuild the final mosaic; dense molecular narratives are summarized into intent axes for readability.}
    \label{fig:qualitative}
\end{figure}

\subsection{Ablation study}\label{sec:ablation}

\noindent\textbf{Component ablations} (Table~\ref{tab:ablation_hr}).
Text-based retrieval outperforms visual-only selection (+0.077 HER2, +0.045 Risk), indicating that caption semantics capture endpoint-relevant cues beyond embedding saliency.
Removing the consistency gate reduces HER2 by 0.043 and Risk by 0.040.
Gate-only (computing $c$ without repair) and Random repair (adding 32 random patches) recover only a fraction of the gap, confirming that deficit-driven re-ranking is the operative component.
All ablations use the same LoRA-tuned diagnosis head; only the retrieval/gating pipeline differs.

\noindent\textbf{Sensitivity and robustness} (Table~\ref{tab:robustness}).
$K{=}16$ balances interpretability and coverage: $K{=}8$ merges distinct cytologic and stromal patterns, while $K{=}32$ adds little gain but requires a larger checklist.
$\tau_c{=}0.5$ is fixed and endpoint-independent; performance is stable from 0.3 to 0.7.
Among $\mathbf{W}$ perturbations, additive noise and entry dropout cause $\le$0.009 degradation; shuffling pathway-to-axis assignments degrades HER2 by 0.033 and Risk by 0.032, supporting that $\mathbf{W}$ encodes informative biological structure rather than arbitrary connectivity.

\subsection{Visualization}\label{sec:viz}

Fig.~\ref{fig:qualitative} shows a held-out case where Stage~1 intent $\mathbf{m}$ highlights proliferation and stromal axes.
Initial retrieval yields $c{=}0.38 < \tau_c$ with necrosis and lymphocytic infiltrate under-represented; after one repair round the rebuilt mosaic reaches $c{=}0.74$.
Globally, the gate triggers on 37\% of test slides, raising $c$ from 0.41 to 0.78 on average and improving HER2 AUC by +0.045 and Risk C-index by +0.038 on the triggered subset. For the remaining 63\% non-triggered cases, the initial zero-shot retrieval was highly aligned (mean $c{=}0.81$), ensuring safety without extra compute overhead.

\section{Conclusion}\label{sec:conclusion}

\noindent\textbf{Limitations.}
Evaluation is limited to internal TCGA-BRCA cross-validation and does not establish robustness across institutions, scanners, stains, or populations.
The breast-cancer-specific 16-axis checklist would need auditing for other tumor types.
VLM captions remain fallible: structured prompts and checklist matching make evidence inspectable, but do not guarantee pathologic correctness or remove same-model bias between captioning and diagnosis.
Inference is also costly at 57\,s/WSI on 2$\times$A100, so faster multi-stage VLM inference is needed before clinical deployment.

\ours{} turns matched multi-omics into a testable morphology hypothesis by mapping omics to an intent vector, retrieving endpoint-relevant regions with TF-IDF captions, and verifying molecular--visual alignment with a cosine consistency gate and deterministic repair.
On TCGA-BRCA (930 WSIs, 5-fold CV), \ours{} achieves state-of-the-art across ER/PR/HER2, subtype, and risk.

\section*{Acknowledgements}
This work was supported by the Open Project Program of the State Key Laboratory of Medical Proteomics (SKLP-O202406), the Emerging Frontiers Cultivation Program of Tianjin University Interdisciplinary Center, the Seed Foundation of Tianjin University, the Internal Research Grants of Macao Polytechnic University (No.~RP/CAI02/2023), and the Science and Technology Development Fund (No.~0177/2023/RIA3).

\section*{Disclosure of Interests}
The authors have no competing interests to declare.

\bibliographystyle{splncs04}
\bibliography{references}

\end{document}